\documentclass[conference]{IEEEtran}
\IEEEoverridecommandlockouts
\usepackage{cite}
\usepackage{amsmath,amssymb,amsfonts}
\usepackage{algorithmic}
\usepackage{graphicx}
\usepackage{epstopdf}
\usepackage{textcomp}
\usepackage{xcolor}
\def\BibTeX{{\rm B\kern-.05em{\sc i\kern-.025em b}\kern-.08em
    T\kern-.1667em\lower.7ex\hbox{E}\kern-.125emX}}
\begin{document}

\title{Compositional Learning in Transformer-Based Human-Object Interaction Detection}

\author{\IEEEauthorblockN{Zikun Zhuang, Ruihao Qian, Chi Xie, Shuang Liang$^\ast$}\thanks{*Corresponding author.}
\IEEEauthorblockA{\textit{School of Software Engineering} \\
\textit{Tongji University}\\
Shanghai, China \\
\{2131488, 2031541, chixie, shuangliang\}@tongji.edu.cn}
}

\maketitle

\begin{abstract}
Human-object interaction (HOI) detection is an important part of understanding human activities and visual scenes. The long-tailed distribution of labeled instances is a primary challenge in HOI detection, promoting research in few-shot and zero-shot learning. Inspired by the combinatorial nature of HOI triplets, some existing approaches adopt the idea of compositional learning, in which object and action features are learned individually and re-composed as new training samples. However, these methods follow the CNN-based two-stage paradigm with limited feature extraction ability, and often rely on auxiliary information for better performance. Without introducing any additional information, we creatively propose a transformer-based framework for compositional HOI learning. Human-object pair representations and interaction representations are re-composed across different HOI instances, which involves richer contextual information and promotes the generalization of knowledge. Experiments show our simple but effective method achieves state-of-the-art performance, especially on rare HOI classes.
\end{abstract}


\section{Introduction}

\label{sec:intro}
Human-object interaction (HOI) detection aims to localize human and object instances in a given image, and recognize interactions of human-object pairs. The task is also formulated as detection of HOI triplets ⟨human, verb, object⟩. Existing HOI detection methods can be divided into two-stage and one-stage methods. Two-stage methods~\cite{b1,b2,b3} sequentially perform two sub-tasks: object detection and interaction classification. Humans and objects are detected and paired as human-object proposals, and the interaction classifier predicts interaction classes of proposals based on visual features of human-object pairs. Diverse auxiliary information, including human-object spatial configuration~\cite{b1}, human pose~\cite{b2} and language priors~\cite{b3}, is introduced to provide cues. One-stage methods eliminate the process of enumerating human-object proposals for higher efficiency. Early one-stage approaches~\cite{b4,b5} detect points or regions of interaction, and perform object detection and interaction classification in parallel. In recent one-stage methods~\cite{b6,b7,b8}, various network architectures based on transformer~\cite{b9} are proposed to perform end-to-end HOI detection.

\begin{figure}[ht]
	\centering
	\includegraphics[width=\columnwidth]{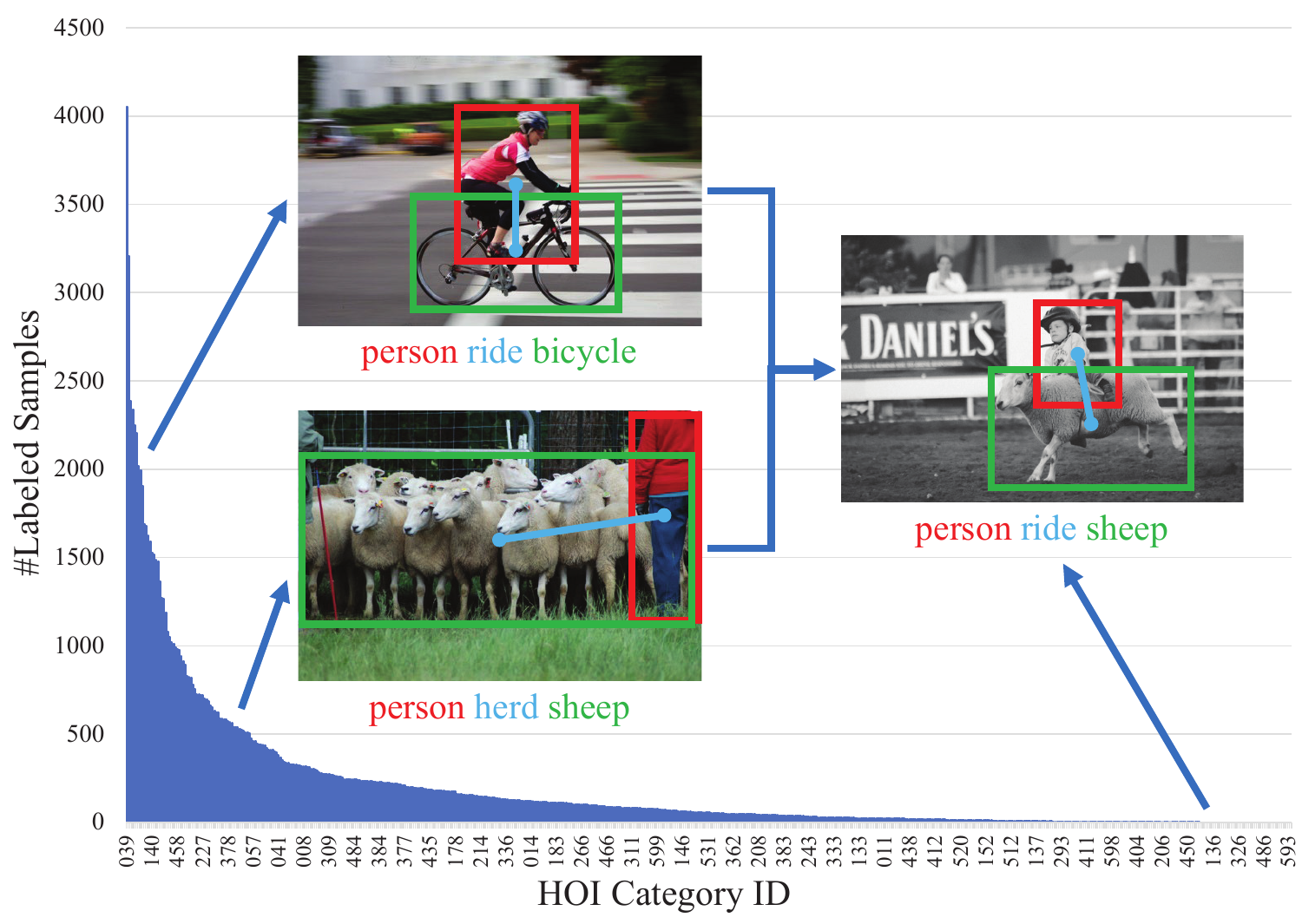}
	\caption{The idea of compositional HOI learning. HOI triplets are decomposed into persons, actions and objects, and these components are learned individually and re-composed as new HOI classes. This helps to generalize knowledge from non-rare classes to rare classes.}
	\label{fig:concept}
\end{figure}

A primary challenge in HOI detection lies in the long-tailed distribution of labels of different categories. In HICO-Det~\cite{b1}, 138 of the total 600 HOI classes have only less than 10 samples, leading to insufficient training and poor detection accuracy on rare classes. To address this problem, few-shot and zero-shot learning methods~\cite{b3,b10,b11,b12} have been proposed and improved detection accuracy of rare and unseen classes. The combinatorial characteristic of HOI triplets naturally inspires the idea of compositional learning~\cite{b12}, in which verbs and objects are learned individually and re-composed to form samples of different HOI classes, as Fig.~\ref{fig:concept} illustrates.

However, most of existing compositional learning methods~\cite{b11,b12,b13,b14} follow the traditional two-stage paradigm. Limited by the feature extraction ability of CNN, the original and re-composed samples contain inadequate semantics to infer HOI classes. Moreover, most of these methods only involve object and action features of their local regions in feature re-composition, causing further loss of global contextual information in re-composed feature samples. As a result, these methods often rely on additional information, such as human-object spatial configuration and word embeddings, for better performance. The simple idea of compositional learning remains rarely seen in transformer-based methods. Exploiting the excellent feature extraction capacity of transformer, we believe re-composition between samples with richer visual semantics can promote more comprehensive understanding of HOIs, without introducing any auxiliary information.

We propose a novel transformer-based framework for compositional HOI learning. Given a pair of input images, our model produces human-object pair representations and interaction representations via two cascade decoders of CDN~\cite{b8}. Human-object pair representations predict human and object bounding boxes and object categories, while interaction representations are concatenated with human-object pair representations to predict action classes. We select the representations corresponding to the best predictions matched with ground truths, and concatenate them across different HOI instances as new interaction samples. Labels of re-composed samples are also re-composed with ground truth labels of original samples. On one hand, visual features extracted by transformer contain richer global context, which is involved in our sample re-composition. This enables our model to better understand human-object interactions without the help of additional information. On the other hand, sample re-composition can not only explicitly generalize knowledge to rare classes, but also implicitly encourage the model to learn more generalized knowledge insensitive to changes of object and action classes, and therefore helps eliminate the long-tailed distribution problem. Our main contributions can be summarized as follows: 

\begin{itemize}
	\item To the best of our knowledge, we are the first to apply compositional learning on a transformer-based HOI detection framework, without introducing any supplementary information.
	\item We re-compose human-object pair representations and interaction representations between different HOI instances as new training samples, which involves more global contextual information and promotes knowledge generalization across HOI classes.
	\item On two benchmark datasets, our method achieves excellent overall performance and state-of-the-art performance on rare HOI classes.
\end{itemize}

\begin{figure*}[ht]
	\centering
	\includegraphics[width=18cm]{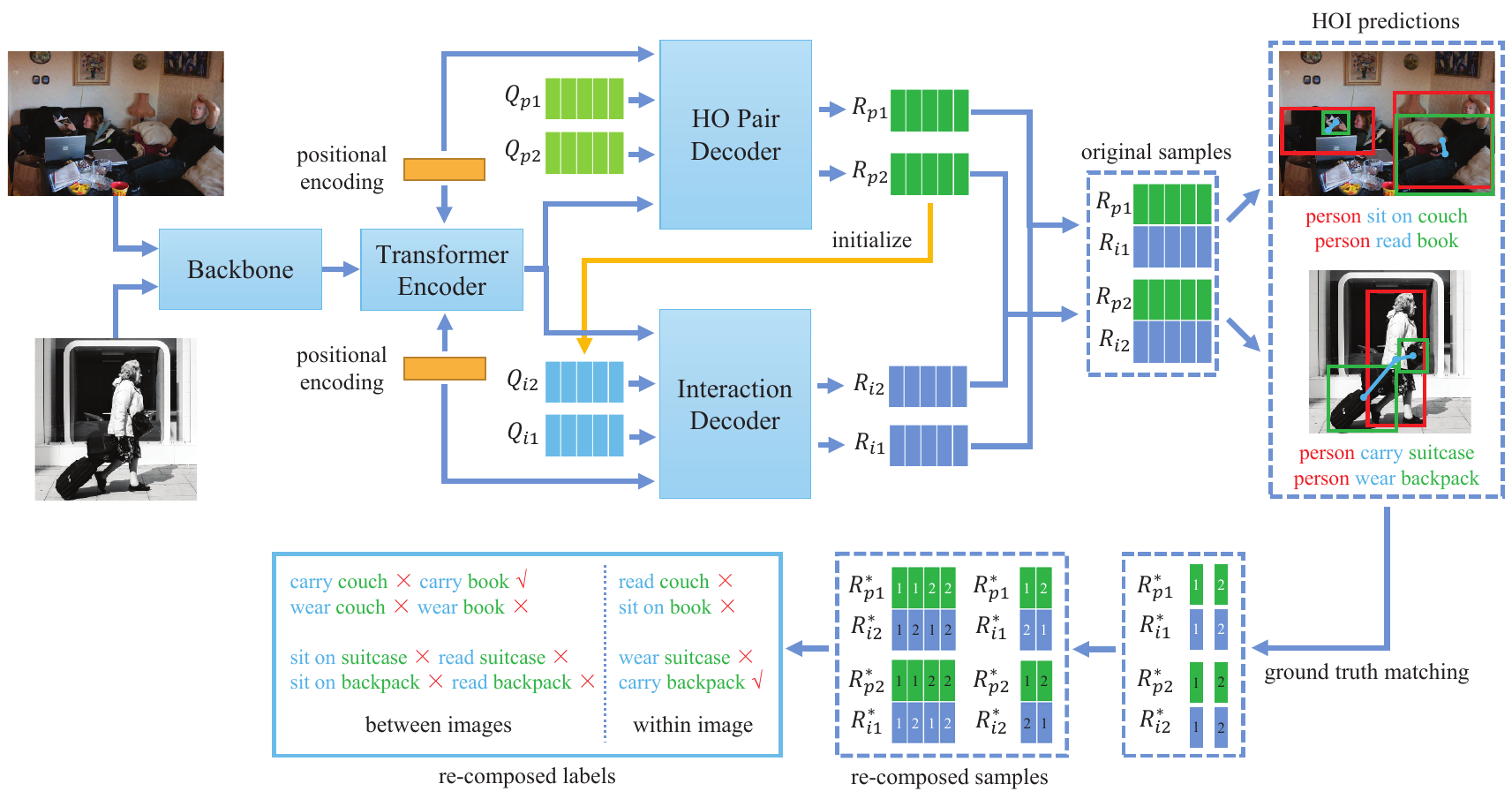}
	\caption{Overview of our method. Taking a pair of images as input, we use the model proposed in~\cite{b8} to transform HO pair queries $Q_p$ and interaction queries $Q_i$ into corresponding representations $R_p$ and $R_i$. Different from~\cite{b8}, we concatenate $R_p$ and $R_i$ to predict action classes. Representations corresponding to predictions that match the ground truths, denoted as $R^\ast_p$ and $R^\ast_i$, are re-composed across different HOI instances, forming new training samples for interaction classification. In re-composed labels, infeasible verb-object compositions beyond dataset definition are removed.}
	\label{fig:overview}
\end{figure*}

\section{Related Work}

\subsection{Few-Shot and Zero-Shot HOI Detection}

Previous few-shot and zero-shot HOI detection methods can be divided into two groups. One group of methods~\cite{b3,b15,b10} adopt vision-language joint learning to augment visual features with linguistic knowledge. Peyre et al.~\cite{b3} learn vision-language embeddings of visual phrases and generate embeddings for unseen triplets via analogies between similar relations. Liu et al.~\cite{b15} construct an knowledge graph with visual-semantic embeddings to encode multi-level relations among objects, actions and HOIs. Liao et al.~\cite{b10} transfers visual-linguistic knowledge from CLIP~\cite{b16} to enhance HOI understanding. The other group of methods re-compose samples to generalize knowledge to rare and unseen HOI classes. Bansal et al.~\cite{b11} propose a functional generalization module where object word embeddings are replaced with those of functionally similar objects during training. Hou et al.~\cite{b12,b13,b14} propose several methods to generate new interaction samples. In VCL~\cite{b12}, they concatenate object and verb features across HOI instances. In FCL~\cite{b13}, they propose to generate fabricated object features containing noise and verb features. In ATL~\cite{b14}, verb features re-defined as affordance features are concatenated with object features from HOI datasets and object detection datasets.

\subsection{Transformer-Based HOI Detection}

Transformer~\cite{b9} has been successfully applied in a diversity of computer vision tasks, \textit{e.g.} DETR~\cite{b17} in object detection, which inspires early transformer-based HOI detection methods~\cite{b6,b7}. In QPIC~\cite{b6}, a transformer encoder aggregates image-wide context, and a transformer decoder transforms a set of queries into embeddings, each directly capturing one human-object pair. In HOTR~\cite{b7}, two decoders respectively generate instance and interaction representations, and each interaction representation associates itself with corresponding humans and objects with HO Pointers. Some works~\cite{b18} exploit deformable attention~\cite{b19} to process multi-scale feature maps for fine-grained HOI detection. Recently, some works~\cite{b8,b20} seek to combine advantages of two-stage and one-stage frameworks. Zhang et al.~\cite{b8} propose two cascade disentangling decoders, each focusing on respective subtasks, \textit{i.e.} object detection and interaction classification. Zhang et al.~\cite{b20} propose a two-stage framework composed of a DETR as object detector and a transformer-based interaction head to pair humans and objects and predict action classes.

\section{Proposed Method}

\subsection{Overview}
Fig.~\ref{fig:overview} illustrates the overall pipeline of our method which is mainly based on CDN~\cite{b8} and VCL~\cite{b12}. Given a pair of input images, the backbone and a shared encoder first extract their global features. Then the human-object pair decoder and the interaction decoder transform a set of learnable queries into corresponding representations. A set of feed-forward networks (FFNs) process human-object pair representations to predict human and object bounding boxes and object categories, and human-object pair representations are concatenated with interaction representations to predict action classes. For compositional learning, we select representations that generate the best predictions matched with ground truths, and concatenate them across different HOI instances to produce new training samples. Labels of re-composed samples are also constructed using labels of original samples. Verb-object compositions that doesn't belong to HOI categories defined by datasets are considered infeasible and removed from re-composed labels.

\subsection{Network Architecture}
\label{sec:arch}

We choose Cascade Disentangling Network (CDN)~\cite{b8} as the baseline model, based on which we implement our compositional learning method because of its effectiveness and code availability. We briefly review its architecture and introduce our modification to it for sample re-composition.

\noindent\textbf{Baseline. }CDN consists of a CNN backbone, a shared encoder and two cascade disentangled decoders, named the Human-Object Pair Decoder (HO-PD) and the interaction decoder, respectively. Given an input image $I\in\mathbb{R}^{H\times W\times C}$, the backbone generates a visual feature map $X_v\in\mathbb{R}^{H'\times W'\times C'}$, which is flattened into the shape of $(H'\times W', C')$ and fed into the encoder along with the positional encoding $E_{pos}\in\mathbb{R}^{(H'\times W')\times C'}$. The encoder further aggregates image-wide context and produces sequenced visual feature vectors $X_s\in\mathbb{R}^{(H'\times W')\times C'}$, denoted as global memory.  

Two decoders take $X_s$, $E_{pos}$ and a set of queries $Q\in\mathbb{R}^{N_q\times C_q}$ as input, apply self-attention on $Q$, conduct multi-head co-attention between $Q$ and $X_s$, and output updated queries denoted as representations $R\in\mathbb{R}^{N_q\times C_q}$. $N_q$ stands for the number of queries. HO-PD $D_p$ decodes features of human-object pairs, while the interaction decoder $D_i$ captures interaction context. $D_p$ and $D_i$ are cascaded by initializing the interaction queries $Q_i$ with HO pair representations $R_p$, so that $Q_p$ are guided by prior knowledge in $R_p$ to learn the corresponding action classes for each HO pair query. The generating process of representations can be described as:
\begin{eqnarray}
	R_p=D_p(Q_p,X_s,E_{pos})
\end{eqnarray}
\begin{eqnarray}
	R_i=D_i(Q_i,X_s,E_{pos})=D_i(R_p,X_s,E_{pos})
\end{eqnarray}
A group of FFNs $F$ process the representations to predict HOI triplets. $R_p$ produce human-object pairs predictions denoted as $Y_p=\{(b^h_n,b^o_n,o_n),n\in\{1,2,...,N_q\}\}$. $b^h_n\in\mathbb{R}^4$, $b^o_n\in\mathbb{R}^4$ and $o_n\in\mathbb{R}^{N_o}$ stand for the human bounding box, the object bounding box and the object class probability distribution, respectively. $N_o$ is the number of object classes. $R_i$ predict action classes of human-object pairs, denoted as $Y_i=\{a_n,n\in\{1,2,...,N_q\}\}$. $a_n\in\mathbb{R}^{N_a}$ is the action class probability distribution, and $N_a$ is the number of action classes. Composed of HO pair predictions and interaction predictions, HOI predictions are denoted as $Y=\{(b^h_n,b^o_n,o_n,a_n),n\in\{1,2,...,N_q\}\}$. The prediction process is formulated as:
\begin{eqnarray}
	Y_k=F_k(R_k),k\in\{p,i\}
\end{eqnarray}

\noindent\textbf{Modified Baseline. }To re-compose original samples and implement our compositional learning method, we make a simple modification to the original baseline. We concatenate HO pair representations with corresponding interaction representations to predict action classes, which can be described as:
\begin{eqnarray}
	Y_i=F_i(R_p\oplus R_i)
\end{eqnarray}
The shape of concatenated representations $R_p\oplus R_i$ is $(N_q, 2\times C_q)$. We denote the model that only concatenates $R_p$ and $R_i$ of the same original samples as the modified baseline.

\subsection{Compositional Learning}

Inspired by Visual Compositional Learning (VCL)~\cite{b12}, we propose our compositional learning method consisting of re-composition of features and re-composition of labels. In our work, sample re-composition is conducted within one pair of randomly selected input images due to hardware limitation, but it's also easier for us to explain our method clearly. If one image contains multiple HOI instances, we also re-compose samples across different instances within the image following the same process.

\noindent\textbf{Feature Re-Composition. }Given a pair of input images $I_1$ and $I_2$, our model first executes the general procedure of HOI detection, producing a set of HOI predictions, each corresponding to an HO pair representation and an interaction representation. If all representations are involved in re-composition, the huge number of re-composed samples will largely increase the computation overhead. Besides, each ground truth is matched with only one prediction for loss calculation during training. It indicates that the majority of predictions are inaccurate, and the corresponding representations contain inadequate semantic information, which barely benefits the understanding of HOIs. Therefore, we only re-compose representations that produce the best-matching predictions, denoted as $R^\ast\in\mathbb{R}^{N_{gt}\times C_q}$. $N_{gt}$ symbolizes the number of ground truths, \textit{i.e.} annotated HOI instances in input images.

HO pair representations $R^\ast_p$ and interaction representations $R^\ast_i$ from different images are concatenated as new feature samples, and fed into the interaction classifier to predict action classes, which can be described as:
\begin{eqnarray}
	Y^\ast_i=F_i(R^\ast_p\otimes R^\ast_i)
\end{eqnarray}
Taking $R^\ast_{p1}$ from $I_1$ and $R^\ast_{i2}$ from $I_2$ as example, every representation in $R^\ast_{p1}$ is concatenated with every representation in $R^\ast_{i2}$. So the shape of re-composed representations $R^\ast_{p1}\otimes R^\ast_{i2}$ is $(N_{gt1}\times N_{gt2}, 2\times C_q)$, and $Y^\ast_{i12}=\{a^\ast_n,n\in\{1,2,...,N_{gt1}\times N_{gt2}\}\}$ contains $N_{gt1}\times N_{gt2}$ verb predictions. Combining human-object predictions of original samples, the HOI predictions of re-composed samples are denoted as $Y^\ast_{12}=\{(b^{h1}_n,b^{o1}_n,o^1_n,a^\ast_n),n\in\{1,2,...,N_{gt1}\times N_{gt2}\}\}$.

In VCL~\cite{b12}, object features from object bounding box regions are re-concatenated with verb features from union regions of human-object pairs. We think this way of re-composition fails to involve enough human features and global context, which are included in HO pair representations. So we believe our feature re-composition method can enhance the generalization of higher-dimensional semantic knowledge.

\noindent\textbf{Label Re-Composition. }Similar to the HOI prediction, an HOI ground truth label consists of the human bounding box, the object bounding box, the object class label and the action class label. We denote the human and object labels as HO pair labels. The ground truth labels of an image are denoted as $\hat Y=\{(\hat b^h_n,\hat b^o_n,\hat o_n,\hat a_n),n\in\{1,2,...,N_{gt}\}\}$. $\hat o_n$ is an one-hot vector and $\hat a_n$ is a multi-hot vector.

Taking HO pair labels from $I_1$ and action labels from $I_2$ as example, we pair each HO pair label in $\hat Y_{p1}$ with every action label in $\hat Y_{i2}$, and check the feasibility of compositions. An object-verb composition that doesn't belong to HOI categories defined by the dataset is considered infeasible, even if it may be rational (\textit{e.g.} “couch” and “wear”, “suitcase” and “sit on” in Fig.~\ref{fig:overview}). By setting the value of infeasible actions in the action label vector to zero, infeasible compositions are removed. If all compositions between the object label and all action labels are infeasible, an all-zero action label vector is kept for the object label instead of removing the whole HOI label, because the HO pair label is still useful for instance detection. The re-composed HOI labels are denoted as $\hat Y^\ast_{12}=\{(\hat b^{h1}_n,\hat b^{o1}_n,\hat o^1_n,\hat a^2_n),n\in\{1,2,...,N_{cp}\}\}$, where $N_{cp}$ stands for the number of kept re-composed labels.


\subsection{Training and Inference}

\noindent\textbf{Training. }For loss calculation, each ground truth finds its best-matching prediction with the Hungarian algorithm used in~\cite{b6}. Following~\cite{b6}, the target loss is the weighted sum of four parts: box regression loss $L_b$, generalized intersection-over-union loss $L_u$~\cite{b21}, object classification loss $L_o$ and action classification loss $L_a$, described as:
\begin{eqnarray}
	L=\lambda_bL_b+\lambda_uL_u+\lambda_oL_o+\lambda_aL_a
\end{eqnarray}
where $\lambda_b$, $\lambda_u$, $\lambda_o$, $\lambda_a$ are hyper-parameters for adjusting weights of each loss. We use the above loss function for both original and re-composed samples. In a mini-batch, the average loss is the weighted sum of the loss of original samples $L_{orig}$ and the loss of re-composed samples $L_{compo}$:
\begin{eqnarray}
	L_{batch}=\rho L_{orig}+(1-\rho)L_{compo}
\end{eqnarray}
where $\rho$ is the hyper-parameter that adjusts weights of original and re-composed samples. To avoid re-composed samples from dominating the HOI learning, we suggest to give a larger weight for original samples, \textit{i.e.} set $\rho>0.5$.

\noindent\textbf{Inference. }We perform standard HOI detection without sample re-composition during inference. Following~\cite{b8}, we generate the $n$-th prediction as $(b^h_n, b^o_n, argmax_k s^{hoi}(k))$. The HOI class score $s^{hoi}_n=s^o_ns^a_n$, where $s^o_n$ and $s^a_n$ are scores of corresponding object and action class, respectively.

\section{Experiments}

\subsection{Datasets and Metrics}

\noindent\textbf{Datasets. }For performance evaluation, we adopt two popular benchmark datasets, V-COCO~\cite{b22} and HICO-Det~\cite{b1}. Both datasets contain 80 object classes defined by MS-COCO~\cite{b23}. V-COCO includes 2,533 images for training, 2,867 images for validation and 4,946 images for testing. Each human instance is annotated with binary labels for 29 action categories. HICO-Det consists of 38,118 images for training and 9,658 images for testing, providing more than 150,000 annotated HOI instances. It contains 117 action classes and explicitly defines 600 HOI categories.

\noindent\textbf{Metrics. }Following~\cite{b1}, we use mean average precision (mAP) as the evaluation metric. An HOI prediction is considered as a true positive when both bounding boxes of human and object have intersection over union (IoU) with a ground truth greater than 0.5, and the predicted HOI label is correct. On HICO-Det we report mAPs under the default setting on three category sets: all 600 classes (Full), 138 classes with less than 10 training samples (Rare) and the other 462 classes (Non-Rare). On V-COCO we report role mAPs under Scenario 1 following its official evaluation setup.

\subsection{Implementation Details}

CDN~\cite{b8} provides PyTorch implementations of three variant architectures: CDN-S (small), CDN-B (base) and CDN-L (large). For quick validation, we implement our method on CDN-S, which consists of a ResNet-50 as backbone, a 6-layer encoder and two 3-layer decoders. Box regression FFNs have 3 linear layers with ReLU, while object and action classfiers are single-layer FFNs. The number of queries $N_q$ is set to 64 for HICO-Det and 100 for V-COCO. The number of channels of each query $C_q$ is set to 256. Following ~\cite{b6}, weight coefficients $\lambda_b$, $\lambda_u$, $\lambda_o$, $\lambda_a$ are set to 2.5, 1, 1 and 1, respectively. Same as~\cite{b8}, we initialize the network with pre-trained parameters of DETR~\cite{b17}, and use AdamW with a weight decay of $10^{-4}$ for optimization. We first train the whole model for 90 epochs with a learning rate of $10^{-4}$, decreasing by 10 times at 60th epoch. Then we fine-tune the cascade decoders together with FFNs for 10 epochs with a learning rate of $10^{-5}$. We also use the same dynamic re-weighting strategy and post-processing of pairwise non-maximal suppression as~\cite{b8} proposes. All experiments are conducted on 2 RTX 2080 Ti GPUs with the batch size of 2.

\begin{table}[t]
	\begin{center}
		\caption{Ablation study of our method on HICO-Det.} \label{tab:ablation-hico}
		\begin{tabular}{|c|c|c|c|}
			\hline
			Method & Full & Rare & Non-Rare
			\\
			\hline
			Baseline & 29.45 & 23.35 & 31.27 \\
			Baseline$^\ast$ & 29.58 & 23.82 & 31.30 \\
			Compo ($\rho=0.5$) & 28.96 & 23.88 & 30.48 \\
			Compo ($\rho=0.75$) & \textbf{30.03} & \textbf{24.62} & \textbf{31.65} \\
			Compo ($\rho=0.9$) & 29.43 & 23.34 & 31.25 \\
			\hline
		\end{tabular}
	\end{center}
\end{table}
\begin{table}[t]
	\begin{center}
		\caption{Ablation study of our method on V-COCO.} \label{tab:ablation-vcoco}
		\begin{tabular}{|c|c|}
			\hline
			Method & $mAP_{role}$ 
			\\
			\hline
			Baseline & 55.98 \\
			Baseline$^\ast$ & 55.84 \\
			Compo ($\rho=0.5$) & 56.74 \\
			Compo ($\rho=0.75$) & 56.48 \\
			Compo ($\rho=0.9$) & \textbf{57.24} \\
			\hline
		\end{tabular}
	\end{center}
\end{table}

\subsection{Ablation Study}

We first validate the effectiveness of our method compared with baseline models. Experimental results are demonstrated in Table~\ref{tab:ablation-hico} and Table~\ref{tab:ablation-vcoco}, where Baseline, Baseline$^\ast$ and Compo denote the original CDN-S, the modified baseline mentioned in Section~\ref{sec:arch} and our sample re-composition method, respectively.

\begin{figure}[ht]
	\centering
	\includegraphics[width=\columnwidth]{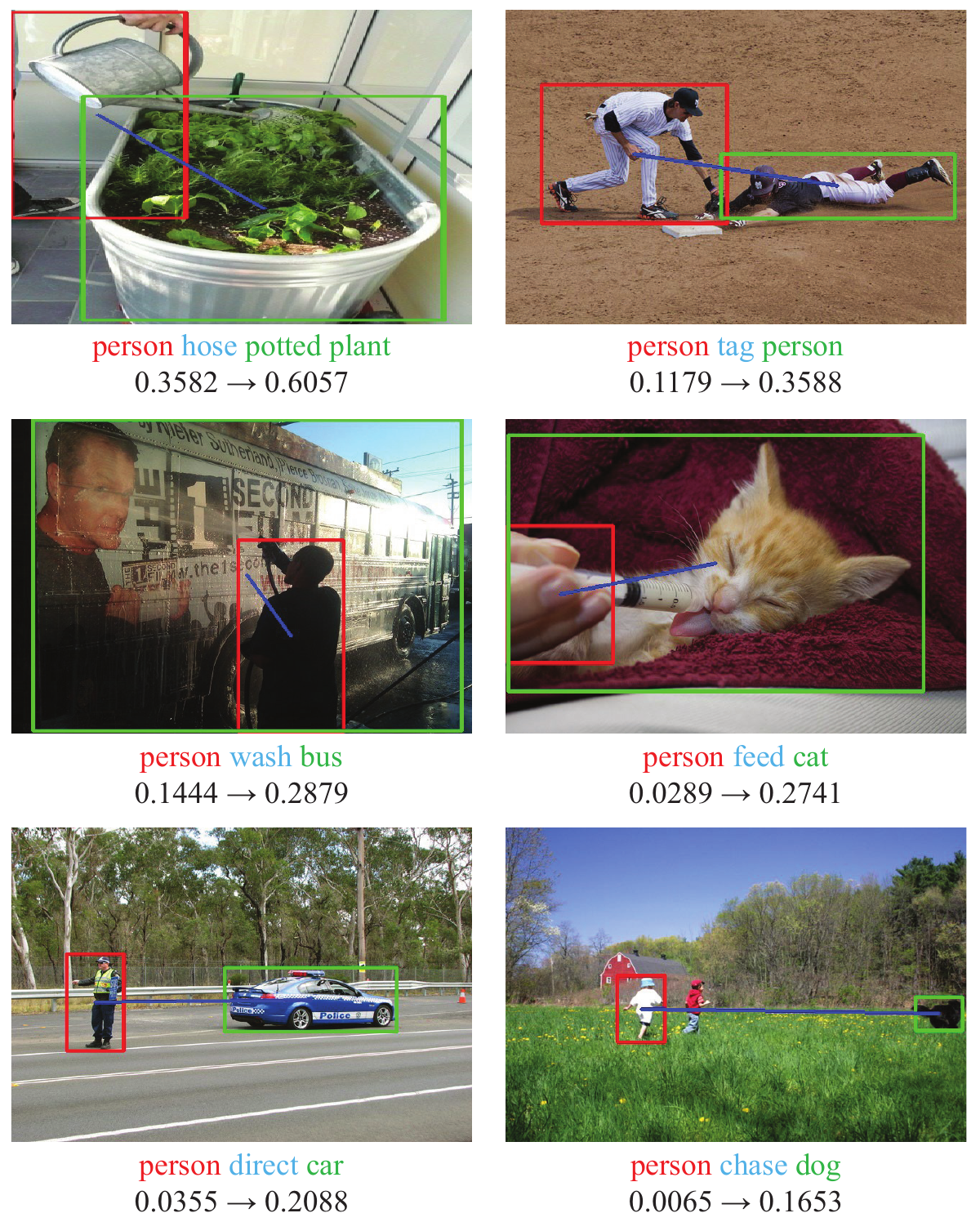}
	\caption{Qualitative results on HICO-Det. We compare HOI triplet scores predicted by Baseline and Compo ($\rho=0.75$). Our method significantly improves detection of rare HOI classes.}
	\label{fig:quali}
\end{figure}

As Table~\ref{tab:ablation-hico} shows, Baseline$^\ast$ which concatenates representations of HO pairs and interactions to predict action classes outperforms the original baseline on HICO-Det, especially improving the mAP on rare classes by a large margin. Applying sample re-composition on the basis of Baseline$^\ast$, our method achieves the best performance with $\rho=0.75$ on three category sets of HICO-Det. Note that our best model significantly increases the mAP on rare classes from 23.35 to 24.62 compared with Baseline. This proves the effectiveness of our method in eliminating the long-tailed distribution problem, which is also shown by qualitative results in Fig.~\ref{fig:quali}.

According to Table~\ref{tab:ablation-vcoco}, our method also performs better compared with baselines on V-COCO, achieving peak performance with $\rho=0.9$. We assume that the difference of best $\rho$ values on two datasets may result from the number of feasible labels. On HICO-Det with a larger range of object and action categories, the re-composition goes beyond defined HOI classes far more easily than on V-COCO, leading to less feasible HOI labels in re-composed samples. Therefore, re-composed samples of HICO-Det contains less knowledge than those of V-COCO, and a slightly larger weight for re-composed samples becomes necessary.

\begin{table}[t]
	\begin{center}
		\caption{Comparison with state-of-the-art methods on HICO-Det.}
		\label{tab:comparison-hico}
		\resizebox{\linewidth}{!}{
			\begin{tabular}{|c|c|c|c|c|}
				\hline
				Method & Backbone & Full & Rare & Non-Rare
				\\
				\hline
				Analogy~\cite{b3} & ResNet-50+FPN & 19.40 & 14.60 & 20.90 \\
				Functional~\cite{b11} & ResNet-101 & 21.96 & 16.43 & 23.62 \\
				VCL~\cite{b12} & ResNet-50 & 19.43 & 16.55 & 20.29 \\
				VCL~\cite{b12} & ResNet-101 & 23.63 & 17.21 & 25.55 \\
				ATL~\cite{b14} & ResNet-101 & 24.50 & 18.53 & 26.28 \\
				FCL~\cite{b13} & ResNet-101 & 24.68 & 20.03 & 26.07 \\
				ConsNet~\cite{b15} & ResNet-50+FPN & 25.94 & 19.35 & 27.91 \\
				HOTR~\cite{b7} & ResNet-50 & 25.10 & 17.34 & 27.42 \\
				QPIC~\cite{b6} & ResNet-50 & 29.07 & 21.85 & 31.23 \\
				\hline
				Compo ($\rho=0.5$) & ResNet-50 & 28.96 & 23.88 & 30.48 \\
				Compo ($\rho=0.75$) & ResNet-50 & \textbf{30.03} & \textbf{24.62} & \textbf{31.65} \\
				Compo ($\rho=0.9$) & ResNet-50 & 29.43 & 23.34 & 31.25 \\
				\hline
			\end{tabular}
		}
	\end{center}
\end{table}
\begin{table}[t]
	\begin{center}
		\caption{Comparison with state-of-the-art methods on V-COCO.} \label{tab:comparison-vcoco}
		\begin{tabular}{|c|c|c|}
			\hline
			Method & Backbone & $mAP_{role}$ 
			\\
			\hline
			VCL~\cite{b12} & ResNet-50 & 48.30 \\
			FCL~\cite{b13} & ResNet-101 & 52.35 \\
			ConsNet~\cite{b15} & ResNet-50+FPN & 53.20 \\
			HOTR~\cite{b7} & ResNet-50 & 55.20 \\
			QPIC~\cite{b6} & ResNet-50 & \textbf{58.80} \\
			\hline
			Compo ($\rho=0.5$) & ResNet-50 & 56.74 \\
			Compo ($\rho=0.75$) & ResNet-50 & 56.48 \\
			Compo ($\rho=0.9$) & ResNet-50 & 57.24 \\
			\hline
		\end{tabular}
	\end{center}
\end{table}

\subsection{Comparison with State-of-the-Art}

We compare our method with existing state-of-the-art few-shot learning methods and transformer-based methods of HOI detection. For fair comparison, we give priority to implementations that adopt Resnet-50 as the backbone. From Table~\ref{tab:comparison-hico} we can see our best model outperforms all listed previous methods on HICO-Det, especially exceeding the state-of-the-art mAP on rare classes by a large margin. As Table~\ref{tab:comparison-vcoco} shows, our best model also achieves competitive performance on V-COCO. We believe our method can outperform state-of-the-art methods if implemented on a larger network architecture, \textit{i.e.} CDN-B or CDN-L.

\section{Conclusion}

We propose a novel transformer-based compositional learning framework for few-shot HOI detection. Human-object pair representations and interaction representations from different HOI instances are re-composed as new training samples. This promotes the transfer of knowledge from non-rare classes to rare classes, encourages the learning of generalized semantics, and helps eliminate the long-tailed distribution problem. Experiments on two benchmark datasets prove our method enhances understanding of HOIs without introducing additional information, and achieves state-of-the-art performance especially on rare categories.

\section*{Acknowledgment}

This work was supported in part by the National Natural Science Foundation of China under Grant 62076183, 61936014 and 61976159, in part by the Natural Science Foundation of Shanghai under Grant 20ZR1473500, in part by the Shanghai Science and Technology Innovation Action Project of under Grant 20511100700 and 22511105300, in part by the Shanghai Municipal Science and Technology Major Project under Grant 2021SHZDZX0100, and in part by the Fundamental Research Funds for the Central Universities. The authors would also like to thank the anonymous reviewers for their careful work and valuable suggestions.

\end{document}